\documentclass[sigconf,nonacm]{acmart}
\usepackage{stfloats}  
\usepackage{balance}   
\usepackage{enumitem}
\usepackage{multirow}   
\usepackage{colortbl}   
\usepackage{booktabs}   
\AtBeginDocument{%
  }

\setcopyright{acmlicensed}
\copyrightyear{2018}
\acmYear{2018}
\acmDOI{XXXXXXX.XXXXXXX}
\acmConference[Conference acronym 'XX]{Make sure to enter the correct
  conference title from your rights confirmation email}{June 03--05,
  2018}{Woodstock, NY}
\acmISBN{978-1-4503-XXXX-X/2018/06}

\begin{document}

\title{Taming a Retrieval Framework to Read Images in Humanlike Manner for Augmenting Generation of MLLMs}


\author{Suyang Xi}
\email{Merwinn520@outlook.com}
\affiliation{%
  \institution{Emory University}
  \city{Atlanta}
  \country{USA}
}

\author{Chenxi Yang}
\affiliation{%
  \institution{University of Electronic Science and Technology of China}
  \city{Chengdu}
  \country{China}
}

\author{Hong Ding}
\affiliation{%
 \institution{University of Illinois Chicago}
 \city{Chicago}
 \country{USA}
}

\author{Yiqing Ni}
\affiliation{%
  \institution{The Hong Kong Polytechnic University}
  \city{Hong Kong}
  \country{China}
}

\author{Catherine C. Liu}
\affiliation{%
  \institution{The Hong Kong Polytechnic University}
  \city{Hong Kong}
  \country{China}
}

\author{Yunhao Liu}
\affiliation{%
  \institution{The Hong Kong Polytechnic University}
  \city{Hong Kong}
  \country{China}
}

\author{Chengqi Zhang}
\affiliation{%
  \institution{The Hong Kong Polytechnic University}
  \city{Hong Kong}
  \country{China}
}

\renewcommand{\shortauthors}{Trovato et al.}

\begin{abstract}

Multimodal large language models (MLLMs) often fail in fine-grained
visual question answering, producing hallucinations about object identities, positions, and relations because textual queries are not
explicitly anchored to visual referents. Retrieval-augmented generation
(RAG) alleviates some errors, but it fails to align with human-like processing at both the retrieval and augmentation levels. Specifically, it focuses only on global-level image information but lacks local detail and limits reasoning about fine-grained interactions. To overcome this limitation, we present \textit{Human-Like Retrieval-Augmented Generation (HuLiRAG)}, a framework that stages multimodal reasoning as a ``what--where--re\-weight'' cascade. Queries are
first anchored to candidate referents via open-vocabulary detection
(\textit{what}), then spatially resolved with SAM-derived masks to recover fine-grained precision (\textit{where}), and adaptively prioritized through the trade-off between local and global alignment (\textit{re\-weight}). Mask-guided fine-tuning further injects spatial evidence into the generation process, transforming grounding from a passive bias into an explicit constraint on answer formulation. Extensive experiments demonstrate that this human-like cascade improves grounding fidelity and factual consistency while reducing hallucinations, advancing multimodal question answering toward trustworthy reasoning.

\end{abstract}

\ccsdesc[500]{Computing methodologies~Artificial intelligence}
\ccsdesc[500]{Information systems~Information retrieval}

\keywords{Multimodal retrieval-augmented generation, Explainable recommendation, Visual grounding}

\maketitle

\section{Introduction}

\begin{figure}[t]
  \includegraphics[width=\columnwidth]{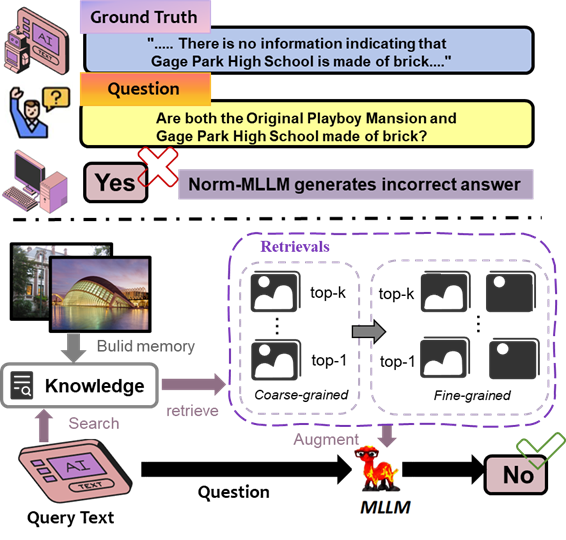}
  \caption{Standard MLLMs struggle with factual VQA due to inadequate perceptual grounding. Our method equips LLMs with the ability to ‘read’ images by dynamically retrieving and aligning semantically relevant visual regions, enabling evidence-based reasoning.
}
  \label{fig:experiments}
\end{figure}

Recent advancements in multimodal LLMs (MLLMs) such as GPT-4o, Gemini, Claude 3, and LLaVA have shown impressive capabilities in a variety of tasks \cite{radford2021learning, alayrac2022flamingo, openai2023gpt4, google2023gemini, anthropic2023claude, liu2023llava}. 
However, these models continue to struggle with fine-grained, fact-based Visual Question Answering (VQA), particularly in scenarios requiring precise, region-specific visual understanding \cite{mensink2023encyclopedic, lin2023fine, lin2024preflmr}. A key limitation of current systems is their inability to dynamically process and validate visual information, relying instead on pre-trained, static knowledge representations \cite{kirillov2023segment, carion2020end, ren2015faster}. This results in two primary challenges: (1) \textit{hallucinations}, where models generate information not grounded in the input data \cite{rawte2023survey, zhou2023detecting, rohrbach2018object}, and (2) \textit{rigid reasoning}, characterized by the lack of flexible, real-time integration of visual input and contextual information, akin to human working memory processes \cite{vaswani2017attention, hudson2018compositional, chen2023pali}. Although current multimodal models have made considerable progress in many areas \cite{liu2024multimodal, zhang2023survey, bang2023multimodal}, these perceptual and reasoning gaps remain a fundamental obstacle in fine-grained VQA tasks.

To address static knowledge limitations in multimodal reasoning, Retrieval-Augmented Generation (RAG) systems augment Multimodal Large Language Models (MLLMs) by dynamically retrieving external evidence\cite{yasunaga2023ra-cm3}. Current approaches can be divided into three paradigms:  
(1) \textit{Efficient Search}: Optimized via Maximum Inner Product Search (MIPS)\cite{guo2020scann} or graph-based ANNs\cite{zhang2024ann} for low-latency retrieval;  
(2) \textit{Modality Specialization}: Leveraging domain-specific embeddings (e.g., CLIP\cite{radford2021clip}, BLIP\cite{li2022blip}) for cross-modal alignment;  
(3) \textit{Adaptive Re-ranking}: Refining results via precision filters\cite{zhang2024h}. However, current retrieval-augmented generation (RAG) systems predominantly operate at the image-level granularity, which proves inadequate for complex multimodal tasks demanding fine-grained understanding~\cite{abootorabi2025ask}. Although some works have begun exploring more flexible retrieval approaches, for example, KURAG ~\cite{zhang2025fine} advances retrieval flexibility through knowledge unit fusion (joint embeddings of text snippets and entity-linked images)~\cite{chen2024limitations}, it still fails to achieve true sub-image-level retrieval (e.g., localized regions or instance-specific patches)~\cite{miller2023dynamic}. This limitation starkly contrasts with human visual cognition, which dynamically integrates multi-granular information, seamlessly shifting between global scenes, local details, and instance features based on contextual demands~\cite{kosslyn2023imagery}.


Fundamentally, existing RAG frameworks exhibit a cognitive-system divergence in compositional reasoning mechanisms. Humans consciously manipulate mental imagery~\cite{kosslyn2023imagery} to compositionally explore cross-modal content across granularities, 
whereas existing RAG approaches tend to entangle signals from multiple modalities into a unified representation space~\cite{zhou2024explainability}, simplifying the problem to facilitate subsequent retrieval and augmentation. However, they lack prior decomposition and analysis of the information sources, making this paradigm prone to rigidity and unable to emulate the nuanced trade-offs and reasoning capabilities humans employ when tackling complex problems.
Cognitively, human retrieval is guided by task-dependent visual saliency~\cite{itti2024task}, while most retrieval systems rely on monolithic similarity metrics, neglecting essential fine-grained visual anchors~\cite{hochstein2023anchors}. This misalignment necessitates rethinking retrieval paradigms—transitioning from rigid deep learning architectures toward neurocognitively-inspired frameworks that emulate human perceptual dynamics~\cite{gershman2024foundations}.

In recent years, vision foundation models~\citep{liu2023grounding, kirillov2023segment} and multimodal foundation models~\citep{sun2024alpha} have advanced rapidly, providing us with tools to process, extract, and analyze image information at the sub-image level. Building on the current powerful foundation models and tools, as well as the above insights into the limitations of existing RAG systems, we introduce \textit{Human-Like Retrieval-Augmented Generation (HuLiRAG)}—a retrieval-augmented generation framework designed to emulate the staged dynamics of human perception. Inspired by how the brain incrementally binds referents, parses spatial structure, and reweights evidence in context, our architecture follows a \emph{what–where–reweight} cascade that explicitly aligns memory, attention, and inference with the compositional flow of visual reasoning. HuLiRAG begins with a \emph{Pre-stage} that employs CLIP~\citep{radford2021learning} to align query semantics with image embeddings, forming a high-recall candidate pool. 
The subsequent \emph{What} module decomposes each query into open-vocabulary phrases describing objects, attributes, and relations to be localized. 
These linguistic cues are grounded by the \emph{Where} module, where GroundingDINO~\citep{liu2023grounding} predicts region proposals refined by SAM~\citep{kirillov2023segment} into high-resolution masks, and Alpha-CLIP~\citep{sun2024alpha} integrates global and regional representations to compute adaptive similarity weights. 
Ultimately, the \emph{Reweight} module calibrates the balance between global and local cues through a lightweight, learnable positive–negative optimization, yielding retrieval that is both semantically precise and spatially grounded. Together, these components transform retrieval from a static preprocessing step into an active perceptual loop—grounded, staged, and contextually responsive. We further consolidate this pipeline with a spatially supervised fine-tuning stage that conditions answer generation on SAM-derived masks, ensuring that reasoning remains anchored in visual evidence. Experiments on WebQA~\citep{yang2022webqa} and MultimodalQA~\citep{talmor2021multimodalqa} demonstrate that region-aware reranking and spatial supervision jointly improve factual consistency and referential grounding, enabling language models to “read” images with human-like perceptual structure.


\section{Related Work}

\textbf{Retrieval-Augmented Language Models (RALMs).}
The paradigm of retrieval-augmented language modeling has evolved significantly from early static retrieval architectures that condition generation on a fixed set of passages~\cite{lewis2020rag, guu2020realm}, to more adaptive frameworks featuring differentiable dense retrievers~\cite{karpukhin2020dpr}, late-interaction encoders for efficient passage scoring~\cite{izacard2022late}, and conditional computation strategies that dynamically adjust retrieval cost based on query complexity~\cite{ainslie2023conditional}. 
Further developments have introduced iterative retrieve–generate loops~\cite{jiang2023iterative}, learned retrieval policies~\cite{borgeaud2023retro}, and memory-augmented transformers that support continual external knowledge access~\cite{wu2022memorizing}. 
While these methods have greatly improved retrieval efficiency and adaptability in text-based settings, they remain inherently text-centric and thus ill-suited for multimodal reasoning. In particular, they struggle to represent visual concepts that demand fine-grained compositional grounding~\cite{hudson2019gqa} and fail to adapt retrieval to query-specific cross-modal signals~\cite{chen2022crossmodal}, both of which are essential for grounded visual question answering and instance-level image–text retrieval.

\begin{figure*}[!t]
\centering
\includegraphics[width=\textwidth]{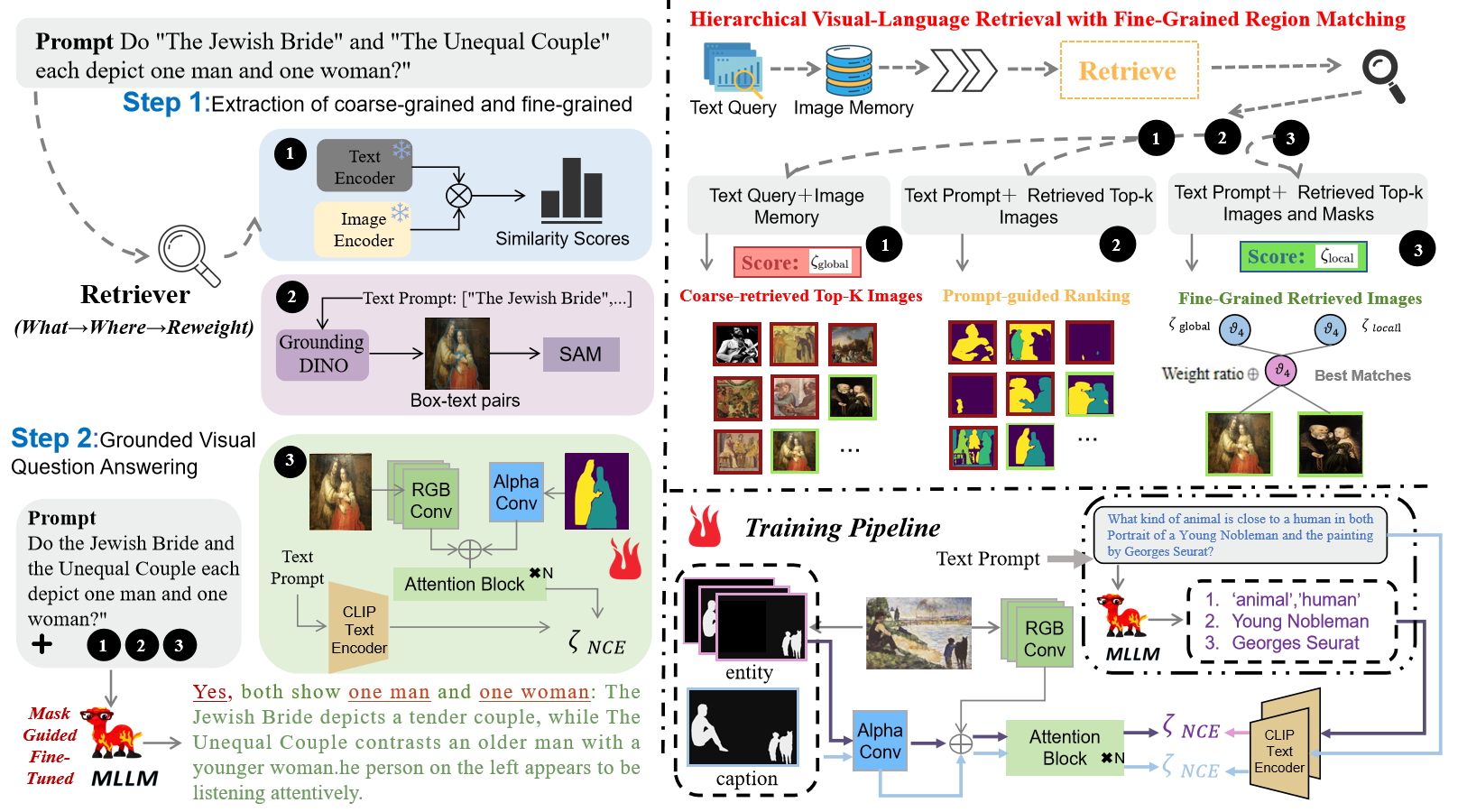}
\caption{Overview of the \textbf{HuLiRAG} framework. Our method implements a human-like staged retrieval pipeline: (1) global feature matching for candidate shortlisting, (2) query-grounded region refinement via detection and segmentation, and (3) adaptive fusion of global and local cues through a learned weighting scheme. The refined evidence feeds into a grounded VQA module, where Alpha-CLIP jointly optimizes global–local alignment.}
\label{fig:framework}
\end{figure*}

\textbf{Multimodal Retrieval-Augmented Generation.}  
Recent advances in retrieval-augmented generation have extended beyond text-only applications into multimodal settings, where external information retrieval supports vision-language tasks such as VQA, captioning, and instruction following~\cite{lewis2020rag, guu2020realm, borgeaud2022retro}. Core architectures differ in their retrieval strategies: some rely solely on language-based retrieval from text corpora, disregarding the visual context~\cite{yang2023multimodal}; others embed image-text pairs into joint latent spaces and retrieve based on vector similarity~\cite{karpukhin2020dpr, xiong2021answering, alayrac2022flamingo}, but these representations often entangle semantic and spatial cues in a manner that lacks interpretability~\cite{zhou2024imagebind}. Multi-stage pipelines aim to improve retrieval precision through coarse-to-fine selection and fusion-based reranking~\cite{asai2020retrieval, izacard2022few, dai2023instructblip}, and recent work has explored symbolic enhancement~\cite{shen2024mmkrag}, multi-hop reranking~\cite{liu2024mr2ag}, and the use of multimodal large language models (MLLMs) themselves as rerankers~\cite{chen2024mllm}. Notably, Kosmos-2~\cite{zhang2023kosmos} adds grounding signals to MLLMs for VQA, but does not directly address staged, region-level reasoning. Despite architectural diversity, most systems treat retrieved content as static evidence—globally conditioned, query-invariant, and aggregated without modeling dynamic, compositional alignment. This design limits the model’s ability to support instance-level reasoning, structured grounding, and flexible adaptation to multimodal query semantics~\cite{kosslyn2003imagery, mao2019counterfactual, zellers2022merlotreserve}. Our approach therefore reframes retrieval as a neurocognitively staged process rather than enlarging MLLMs themselves.

\section{Method}

Humans do not retrive visual informations by processing an image as a single whole. 
Instead, perception unfolds in stages: a global impression captures the scene, attention narrows to likely regions, and localized cues are weighted before forming a response. 
This routine of coarse recognition, targeted inspection, and evidence integration underpins our \textit{Human-Like Retrieval-Augmented Generation (HuLiRAG)} framework. 
As shown in Figure~\ref{fig:framework}, HuLiRAG organizes retrieval into a staged pipeline consisting of a pre-stage coarse candidate retrieval (\S3.1), which is a typical naive retrieval approach, followed by three modules: \textit{What} for query-to-region decomposition (\S3.2), \textit{Where} for visual grounding (\S3.3), and \textit{Reweight} for adaptive evidence allocation (\S3.4). 
On top of this staged retrieval, we introduce \textit{Spatially-Aware Fine-Tuning with Mask-Guided Supervision} (\S3.5), which grounds generation directly in localized evidence. Working in concert, these modules emulate human-like perceptual staging by dynamically binding symbolic queries to spatially grounded visual evidence through a ``what--where--reweight--answer'' process, enabling compositional and context-sensitive reasoning.

\subsection{Pre-stage: Coarse Candidate Retrieval}
\label{sec:clipretrieval}
Before entering the human-like reasoning cascade (\textit{What--Where--Reweight}), 
HuLiRAG begins with a \textit{coarse candidate retrieval} step that prunes the visual search space. 

We begin retrieval by selecting a compact candidate set that coarsely aligns with the query semantics. 
Given a query $q$ and an image corpus $\mathcal{I} = \{I_1, \dots, I_N\}$, 
we adopt frozen CLIP~\citep{radford2021learning} as a dual encoder to compute 
the text embedding $q_i = E_{\text{text}}(q)$ 
and the image embeddings $v_i = E_{\text{image}}(I_i)$. 
Similarity is computed via cosine similarity:
\begin{equation}
s^{\mathrm{clip}}_j = \frac{q_i^\top v_j}{\|q_i\|\cdot\|v_j\|}.
\end{equation}

All image embeddings are pre-indexed with FAISS~\citep{johnson2019billion} for scalable retrieval. 
At inference, we obtain the top-$K$ candidates by:
\begin{equation}
\mathcal{I}_{\text{top}} = \operatorname{TopK}_{I_i \in \mathcal{I}}\, s^{\mathrm{clip}}_i.
\end{equation}

In this way, we prune the search space from $N$ to $K$, 
while deferring precise grounding and alignment to subsequent modules.






\subsection{What: Query-to-Region Decomposition}
\label{sec:where-stage}

The second stage of our framework specifies \textit{what} to look for, 
mirroring the human ability to identify relevant concepts and entities in a question before searching for their spatial locations. 
Instead of encoding the entire query as a single vector, 
we decompose it into a set of semantically grounded phrases that delineate the key objects, attributes, and relations to be localized in the next stage. Each query $q$ is first decomposed into an ordered sequence of open-vocabulary phrases 
$\mathbf{n}(q) = [n_1, \allowbreak \ldots, \allowbreak n_k]$
, 
where each $n_i$ denotes a minimal noun phrase with optional modifiers 
(e.g., ``man in blue'', ``leftmost sign'', ``two cones''). Coreference resolution consolidates repeated mentions, while numerical and spatial cues are explicitly retained, 
ensuring that $\mathbf{n}(q)$ preserves the fine-grained constraints necessary for subsequent localization.
To enhance robustness, phrase construction is regularized to mitigate redundancy and over-fragmentation. 
We define the lexical representation of a phrase as the lemmatized set of its content words 
(nouns, adjectives, numerals, and spatial modifiers), extracted via dependency arcs and NP chunking with spaCy~\citep{honnibal2017spacy}. 
Phrases with substantial lexical overlap are merged:
\begin{equation}
n_i =
\begin{cases}
n_i, & \text{if } \dfrac{|V(n_i)\cap V(n_j)|}{|V(n_i)\cup V(n_j)|} \le 0.7, \\[6pt]
\text{merge}(n_i,n_j), & \text{otherwise},
\end{cases}
\label{eq:phrase_merge}
\end{equation}
where $V(\cdot)$ denotes the lexical set.  
In addition, repeated mentions are disambiguated by indexing, yielding $n^{(1)}, n^{(2)}, \ldots$, so that each textual reference maps uniquely to a visual region.

\subsection{Where: From Linguistic Decomposition to Visual Grounding}

The second stage of HuLiRAG determines \textit{where} to look and how much to attend within the scene. 
It bridges linguistic structure and spatial reasoning by grounding textual phrases into visual regions and adaptively allocating their evidential strength. 
This stage embodies the human-like transition from identifying what entities are described to understanding where they appear and how relevant they are to the question.

\subsubsection{Phrase-to-Region Grounding.}
Each surviving phrase $n_k$ is grounded in the candidate image $I_j$ through a two-step process. 
GroundingDINO~\citep{liu2023grounding} first predicts bounding boxes conditioned on $n_k$, and we retain only detections with confidence above $0.3$, a setting stricter than the default $0.25$ to suppress low-quality matches. 
The Segment Anything Model (SAM)~\citep{kirillov2023segment} then refines each box $b_k$ into a binary mask $m_k$ using a mask threshold of $0.5$, producing sharper region boundaries while maintaining robustness. 
By associating each $n_k$ with its $m_k$, we establish explicit phrase-to-region alignments with both semantic fidelity and spatial precision.  

The final RGBA patch is constructed as $\tilde{I}_k = I_j \odot m_k$, which preserves the localized structure of the entity while eliminating irrelevant background. 
This yields a set of clean, phrase-specific patches $\{\tilde{I}_k\}$ that serve as reliable inputs for the adaptive reweighting stage.

\subsubsection{Adaptive Evidence Integration}

Within the \textit{where} stage, the model further integrates the grounded regions by estimating their relative evidential strength. Pixel-level normalization and semantic feature weighting ensure that salient regions are emphasized while redundant or noisy cues are suppressed, yielding a spatially coherent representation aligned with the query semantics.

\paragraph{Inference.}
We begin by deriving region weights $\alpha_{jk}$ from soft mask assignments:
\begin{equation}
\alpha_{jk} = \frac{1}{|\Omega(I_j)|} \sum_{p \in \Omega(I_j)} \frac{m_k(p)}{\sum_{l=1}^{T_j} m_l(p) + \epsilon},
\label{eq:alpha_soft}
\end{equation}
where $p$ indexes pixels in the image domain $\Omega(I_j)$, $m_k(p)$ is the binary membership of pixel $p$ in mask $m_k$, and $\epsilon$ avoids division by zero. This formulation enforces a probabilistic partition of the image: overlapping regions share pixels fractionally, and the resulting $\alpha_{jk}$ values sum to unity across regions.

Each masked region $\tilde{I}k$ is then encoded by Alpha-CLIP into a region embedding $r{jk}$, while the query $q_m$ is independently mapped to a text embedding. The overall local relevance of image $I_j$ is computed as a weighted aggregation:
\begin{equation}
s^{\mathrm{local}}_j = \sum_{k=1}^{T_j} \alpha_{jk}\,\cos(q_m, r_{jk}),
\label{eq:local_score}
\end{equation}

where $T_j$ denotes the total number of regions. This inference procedure explicitly links textual phrases to their visual counterparts, while attenuating noisy or overlapping evidence. By construction, reweighting balances fidelity and robustness: salient regions are amplified, redundant regions are down-weighted, and alignment is achieved through a principled fusion of localized evidence.



\paragraph{Training and adaptation.}
We finetune Alpha-CLIP with a single objective that couples global and regional supervision.
For each image $I_j$ with question $q$ and region masks $m_k$ linked to phrases $n_k$ extracted from $q$, we build two example types:
(i) global pairs $(I_j, q)$ to preserve holistic semantics;
(ii) regional pairs $(\tilde{I}_k, n_k)$ or $(\tilde{I}_k, q)$ for fine-grained grounding. 
Formally, the training objective simply sums global and regional contrastive losses:
\begin{equation}
\mathcal{L} = \mathcal{L}_{\mathrm{NCE}}(I_j, q) 
+ \frac{1}{T_j} \sum_{k=1}^{T_j} \mathcal{L}_{\mathrm{NCE}}(\tilde{I}_k, n_k),
\label{eq:training_loss}
\end{equation}
where $\mathcal{L}_{\mathrm{NCE}}$ is the InfoNCE loss applied to both holistic and phrase-grounded pairs. 
All inputs are resized to a common resolution and tokenized as concatenations of either the phrase or the full query. 
To accept RGBA inputs, Alpha-CLIP augments the CLIP image encoder with an Alpha Conv branch in the ViT stem; this branch is zero-initialized to keep the RGB path unchanged at warm-up. 
The text encoder is frozen, deeper transformer layers use a smaller learning rate, and we use hybrid sampling that replaces 10\% of regional pairs with full-image queries ($\alpha=1$) to maintain global understanding.

\subsection{Reweight: Adaptive Balance via Positive Negative Sample Pair}
\label{sec:reranking}

CLIP-based joint image–text embeddings offer strong recall but suffer from \emph{semantic myopia}, capturing only gist-level similarity while blurring fine-grained referents and limiting performance on queries that require localized grounding. At the same time, focusing exclusively on local details overlooks the global context that arises from the interaction among multiple regions. Human perception avoids this limitation by alternating between a holistic understanding of the scene and selective attention to candidate regions.

It is necessary to strike a balance and effectively integrate image-level and subimage-level information. In fact, there are two straightforward strategies to combine global similarity $S_{\text{global}}$ and local similarity $S_{\text{local}}$: addition and multiplication. 
\begin{equation}
S_{\text{Add}} = S_{\text{global}} + S_{\text{local}}, \quad 
S_{\text{Multi}} = S_{\text{global}} \cdot S_{\text{local}}.
\end{equation}

However, both approaches are entirely static—they apply the same fusion rule universally across all retrieval domains and datasets. In reality, different retrieval tasks and datasets inherently exhibit distinct biases. For instance, in medical image retrieval, local similarity is often more critical than global similarity, as pathological lesions are typically small and the organs of interest usually occupy only a small portion of the entire image.



To enhance domain generalization and prevent overfitting to a fixed retrieval bias, we introduce a lightweight calibration stage that adaptively balances global and local similarity after fine-tuning. Instead of relying on a static fusion rule, the final similarity is parameterized as
\begin{equation}
    S_{\text{reweight}} = W_g \cdot S_{\text{global}} + W_l \cdot S_{\text{local}} + B,
\end{equation}
where $W_g$, $W_l$, and $B$ are learnable scalar parameters that control the relative contribution of global and local cues.

Since $S_{\text{reweight}}$ lacks explicit ground-truth supervision, we employ a positive–negative contrastive formulation to optimize these parameters in a self-calibrating manner. For each query, its ground-truth (GT) image is treated as a positive sample, while a hard negative is drawn from the top five non-GT candidates according to the initial retrieval ranking based on global similarity. The training objective encourages the model to assign a similarity score close to 1 for positive pairs and near 0 for negative pairs, which we implement using a mean squared error (MSE) loss defined as
\begin{equation}
    \mathcal{L}_{\text{reweight}} = (1 - S_{\text{reweight}}^{+})^2 + (S_{\text{reweight}}^{-})^2,
\end{equation}
where $S_{\text{reweight}}^{+}$ and $S_{\text{reweight}}^{-}$ denote the reweighted similarity of positive and negative pairs, respectively. This lightweight optimization updates only three scalar parameters ($W_g$, $W_l$, and $B$) while leaving the embedding space intact, enabling HuLiRAG to automatically calibrate the weighting of global and local evidence across heterogeneous retrieval domains.

\subsection{Spatially-Aware Fine-Tuning with Mask-Guided Supervision}
\label{sec:finetuning}

Although dual-scale reranking strengthens retrieval by combining global and local evidence, the answer generation stage itself remains unconstrained, and the spatial information obtained during the retrieval stage cannot be effectively utilized in the generation stage.
As a result, multimodal models may still produce responses that are semantically plausible yet spatially ungrounded, reflecting a persistent gap between retrieval accuracy and reasoning faithfulness. 
To close this gap, we introduce a spatially supervised fine-tuning stage that directly anchors the generation process to localized visual evidence. 
Concretely, we condition the model on both the full image $I_j$ and its masked counterpart $\tilde{I}_k = I_j \odot m_k$, where $m_k$ is the SAM-derived mask obtained from the \textit{where} module of HuLiRAG. 
Let $A^{*}$ denote the ground-truth answer. 
To encourage the generator to utilize both global and localized cues, we randomly drop either $I_j$ or $\tilde{I}_k$ during training, preventing collapse into global-only reasoning. 
The optimization follows the standard VQA objective:
\begin{equation}
\mathcal{L}_{vqa} = - \log p\!\left(A^{*} \mid q,\, I_j,\, \tilde{I}_k \right),
\end{equation}
and can be further regularized by enforcing prediction consistency between the two visual contexts, 
\begin{equation}
\mathcal{L}_{cons} = \left\| p(A \mid q, I_j) - p(A \mid q, \tilde{I}_k) \right\|^2,
\end{equation}
to align predictions across global and masked inputs. Here, $p(A \mid q, I_j)$ and $p(A \mid q, I_j \odot m_k)$ denote the predicted probability distributions over the answer space conditioned on the full image $I_j$ or its masked counterpart $\tilde{I}_k$, respectively, while $\|\cdot\|^2$ denotes the squared Euclidean distance between these two distributions. This encourages the model to remain consistent whether reasoning relies on the full scene or localized evidence. The training objective is defined as $\mathcal{L} = \mathcal{L}{vqa} + \mathcal{L}{cons}$, directly coupling answer supervision with consistency regularization. By leveraging retrieval-derived masks as anchors, the loss unifies evidence selection and prediction, reinforcing semantic faithfulness, spatial grounding, and overall coherence of the HuLiRAG pipeline.


It's strightforward that during inference, we simultaneously input the full image $I_j$ and the masked image $\tilde{I}_k$ to enhance the model's answer accuracy.

\definecolor{slategray}{RGB}{225,230,235}

\begin{table*}[t]
\centering
\small
\caption{
Recall performance (R@k) on MMQA and WebQA. 
“Backbone (Vanilla)” denotes raw retrieval with the base retriever presented in Section 3.1, 
while “+HuLiRAG-Ret” denotes the same retriever augmented with our reranking module (AlphaCLIP included in the module is fine-tuned.) presented in Section 3.2-3.4. 
Note: this table reports retrieval-only results, without the generation stage.
}
\label{tab:alpha_clip_reranking}
\resizebox{\textwidth}{!}{%
\begin{tabular}{l|ccc|ccc||ccc|ccc}
\toprule
\multirow{3}{*}{\textbf{Retriever Backbone}} &
\multicolumn{6}{c||}{\textbf{MMQA}} &
\multicolumn{6}{c}{\textbf{WebQA}} \\
\cmidrule(lr){2-7} \cmidrule(lr){8-13}
& \multicolumn{3}{c|}{Vanilla} & \multicolumn{3}{c||}{+HuLiRAG-Ret} &
  \multicolumn{3}{c|}{Vanilla} & \multicolumn{3}{c}{+HuLiRAG-Ret} \\
\cmidrule(lr){2-4} \cmidrule(lr){5-7} \cmidrule(lr){8-10} \cmidrule(lr){11-13}
& R@1 & R@5 & R@10 & R@1 & R@5 & R@10 & R@2 & R@5 & R@10 & R@2 & R@5 & R@10 \\
\midrule
Clip-ViT-L/14@336px     & 79.13 & 88.26 & 90.97 & 87.57 & 93.63 & 96.95 & 58.37 & 73.52 & 86.74 & 73.41 & 82.80 & 88.26 \\
\rowcolor{slategray}
\quad -Finetuned           & 77.66 & 88.17 & 90.21 & 86.71 & 93.30 & 96.38 & 57.76 & 74.77 & 85.70 & 71.38 & 83.23 & 86.64 \\
AlphaClip-(Full-mask)     & 45.35 & 66.09 & 75.22 & 54.52 & 73.39 & 82.97 & 44.33 & 71.51 & 80.14 & 52.05 & 79.95 & 83.57 \\
\rowcolor{slategray}
\quad -Finetuned           & 49.86 & 72.48 & 80.19 & 59.20 & 79.35 & 86.83 & 48.28 & 73.64 & 83.53 & 57.30 & 82.51 & 87.03 \\

Vis-BGE-base               & 46.52 & 70.87 & 78.26 & 75.25 & 87.03 & 90.14 & 29.15 & 48.23 & 59.46 & 53.22 & 59.31 & 66.71 \\
Vis-BGE-m3                 &  41.30 & 64.78 & 69.57 & 72.96 & 87.91 & 92.36 & 26.84 & 46.56 & 57.51 & 50.76 & 57.24 & 64.97 \\
InternVL-C                 & 78.22 & 90.61 & 94.04 & 88.56 & 94.78 & 97.09 & 65.22 & 81.93 & 88.31 & 83.25 & 89.89 & 91.57 \\
InternVL-G                 & 78.91 & 90.52 & 94.83 & 87.32 & 94.41 & 97.24 & 65.17 & 80.52 & 88.33 & 81.85 & 87.62 & 90.98 \\
\bottomrule
\end{tabular}
}
\end{table*}

\definecolor{slategray}{RGB}{225,230,235}

\definecolor{slategray}{RGB}{225,230,235} 
\definecolor{slategray}{RGB}{225,230,235}

\begin{table}[t]
\centering
\small

\caption{
Retrieval performance of HuLiRAG-Ret across global, local, and fused similarity strategies. 
The "Global" and "Local" levels represent retrieval using only holistic or region-aware similarity without cross-scale fusion, 
whereas the other strategies fuse the two through addition, multiplication, or a learnable reweighting mechanism.
}

\label{tab:rerank-fixed}
\setlength{\tabcolsep}{3.8pt}
\renewcommand{\arraystretch}{1.1}
\begin{tabular}{l|ccc}
\toprule
\textbf{Strategy} & \textbf{MMQA R@1} & \textbf{WebQA R@2} & \textbf{WebQA R@10} \\
\midrule
Global level      & 77.66 & 57.76 & 85.70 \\
Local level       & 83.83 & 67.32 & 86.14 \\
\midrule
Addition          & 86.04 & 70.45 & 86.33 \\
Multiply          & 86.19 & 70.50 & 86.38 \\
Learnable Reweight& 86.71 & 71.38 & 86.64 \\
\bottomrule
\end{tabular}
\end{table}

\section{Experiments}
\label{sec:experiments}
\subsection{Datasets}
\textbf{Experimental Setup.}We evaluate our framework on two established benchmarks that stress different aspects of multimodal reasoning: WebQA~\citep{yang2022webqa} and MultimodalQA~\citep{talmor2021multimodalqa}. WebQA comprises over 43,000 real-world image-question-answer triplets harvested from the web, covering diverse domains with long-tail object distributions and noisy visual-textual alignment. In contrast, MultimodalQA features 29,918 questions grounded in complex tabular visual contexts, requiring multi-hop reasoning over structured spatial layouts.

For evaluation, we adopt two complementary protocols. 
In the retrieval stage, we measure performance with Recall@K ($K{=}1,5,10$) \citep{radford2021learning,yang2022webqa}, 
which assesses whether the ground-truth image is successfully ranked within the top-$K$ candidates. 
For VQA, we adopt Exact Match (EM) on MultimodalQA, token-level F1 on WebQA, 
and additionally employ the LLM-as-a-Judge protocol as a complementary assessment of answer quality. 
These combined metrics provide a holistic evaluation of retrieval precision and reasoning fidelity.

\textbf{Implementation Details.}
All experiments are conducted on a cluster with 8×NVIDIA H20 GPUs, using FP16 mixed-precision training. For MultimodalQA~\citep{talmor2021multimodalqa}, the official test split does not release answer labels; following prior work~\citep{chen2024mllm}, we therefore use the development set as the evaluation benchmark. For WebQA~\citep{yang2022webqa}, which similarly lacks an annotated test set, we adopt the common 8:1:1 random split into training/validation/test subsets, consistent with recent practice~\citep{dai2023invl,zhang2023multimodal}. All images are resized to 336×336 pixels. During inference, the latency of CLIP-based global retrieval is 1.4 seconds per sample, and dual-scale reranking latency is 4.9 seconds per sample.  

We implement a two-path RAG pipeline with global retrieval using CLIP-ViT-L/14@336px~\citep{radford2021learning} and local alignment via Alpha-CLIP with an RGBA branch. Training employs InfoNCE with AdamW (weight decay 0.01), cosine-annealed learning rates, large batches (1024 for CLIP, 512 for Alpha-CLIP), and mixed precision; early stopping is triggered by validation R@1 (patience = 5). At inference, CLIP retrieves the top-20 candidates, which are reranked by Alpha-CLIP and reduced to top-N (N = 2 for WebQA, N = 1 for MultimodalQA). Across this pipeline, we benchmark InternVL and LLaVA backbones~\citep{zhu2025internvl3,wang2025internvl35,liu2024llava} under three regimes: frozen, standard finetuning, and our Mask-Guided Finetuning, demonstrating consistent VQA gains across model scales.

\definecolor{slategray}{RGB}{225,230,235}
\begin{table*}[t]
\centering
\small
\caption{
EM (\%) on MultimodalQA and F1 (\%) on WebQA across inference settings. 
Text-only: question only; CLIP$\rightarrow$VQA: retrieval+VQA; 
HuLiRAG: cascade method; Ground-truth Image (oracle setting with gold-standard image directly given). 
Symbols: $\circ$ Zero-shot, $\blacktriangle$ Finetuned, $\blacksquare$ Mask-Guided FT.
}
\label{tab:mq-webqa-final}
\resizebox{\textwidth}{!}{%
\begin{tabular}{l|l|cccc|ccc}
\toprule
\multirow{2}{*}{\textbf{Model}} & \multirow{2}{*}{\textbf{Regime}} 
& \multicolumn{4}{c|}{\textbf{MultimodalQA (EM, \%)}} 
& \multicolumn{3}{c}{\textbf{WebQA (F1, \%)}} \\
\cmidrule(lr){3-6} \cmidrule(lr){7-9}
& & Text-only & CLIP$\rightarrow$VQA & HuLiRAG & GT Image 
  & CLIP$\rightarrow$VQA & HuLiRAG & GT Image \\
\midrule
\multirow{3}{*}{InternVL-1B}
& $\circ$ Zero-shot & 12.17 & 30.00 & 41.14 & 50.56 & 39.32 & 43.79 & 45.85 \\
& $\blacktriangle$ Finetuned & 21.74 & 55.22 & 57.46 & 61.93 & 71.94 & 72.87 & 75.05 \\
& \cellcolor{slategray}$\blacksquare$ Mask-Guided FT 
  & \cellcolor{slategray}22.81 & \cellcolor{slategray}55.73 & \cellcolor{slategray}59.35 & \cellcolor{slategray}63.84 
  & \cellcolor{slategray}72.82 & \cellcolor{slategray}74.21 & \cellcolor{slategray}76.79 \\
\addlinespace
\multirow{3}{*}{InternVL-2B}
& $\circ$ Zero-shot & 17.39 & 47.39 & 50.37 & 52.29 & 43.75 & 49.76 & 51.22 \\
& $\blacktriangle$ Finetuned & 24.78 & 57.39 & 61.47 & 64.39 & 72.81 & 74.88 & 76.95 \\
& \cellcolor{slategray}$\blacksquare$ Mask-Guided FT 
  & \cellcolor{slategray}26.03 & \cellcolor{slategray}58.66 & \cellcolor{slategray}63.16 & \cellcolor{slategray}66.26 
  & \cellcolor{slategray}74.15 & \cellcolor{slategray}76.27 & \cellcolor{slategray}78.53 \\
\addlinespace
\multirow{3}{*}{InternVL-4B}
& $\circ$ Zero-shot & 29.13 & 50.00 & 52.93 & 55.96 & 50.21 & 53.58 & 56.94 \\
& $\blacktriangle$ Finetuned & 32.17 & 61.74 & 65.17 & 69.27 & 76.82 & 77.49 & 78.07 \\
& \cellcolor{slategray}$\blacksquare$ Mask-Guided FT 
  & \cellcolor{slategray}34.66 & \cellcolor{slategray}63.11 & \cellcolor{slategray}67.52 & \cellcolor{slategray}72.35 
  & \cellcolor{slategray}77.75 & \cellcolor{slategray}79.45 & \cellcolor{slategray}80.20 \\
\addlinespace
\multirow{3}{*}{InternVL-8B}
& $\circ$ Zero-shot & 34.35 & 53.48 & 55.96 & 58.72 & 44.13 & 47.77 & 51.28 \\
& $\blacktriangle$ Finetuned & 36.61 & 67.10 & 71.04 & 75.97 & 74.34 & 75.89 & 77.08 \\
& \cellcolor{slategray}$\blacksquare$ Mask-Guided FT 
  & \cellcolor{slategray}39.43 & \cellcolor{slategray}69.57 & \cellcolor{slategray}74.31 & \cellcolor{slategray}79.88 
  & \cellcolor{slategray}75.72 & \cellcolor{slategray}77.38 & \cellcolor{slategray}78.84 \\
\bottomrule
\end{tabular}
}
\end{table*}

\subsection{Main Results}

\textbf{Fine-grained regional alignment via HuLiRAG delivers consistent and significant retrieval efficacy enhancements.} As shown in Table~\ref{tab:alpha_clip_reranking}, the integration of HuLiRAG substantially improves reranking performance beyond vanilla retrieval across MMQA and WebQA benchmarks. For the widely adopted CLIP-ViT-L/14@\allowbreak336px backbone, MMQA R@1 increases from 79.13\% to 87.57\%, while WebQA R@2 improves from 58.37\% to 73.41\%. Interestingly, the finetuned variant performs slightly worse than the vanilla backbone, suggesting that task-specific adaptation may aggravate overfitting to Internet-scale pretraining data and thus reduce cross-dataset generalization, leaving HuLiRAG to partly restore the gap. In contrast, AlphaCLIP under a full mask regime, where the entire image is uniformly treated as foreground without selective focus, starts from much weaker baselines (45.35\% R@1 on MMQA, 44.33\% R@2 on WebQA), yet still shows strong recovery once paired with HuLiRAG. Beyond CLIP style encoders, the framework consistently enhances Vis-BGE and InternVL families, with Vis-BGE-base gaining +28.73 on MMQA R@1 and +24.07 on WebQA R@2, and InternVL-C advancing to 97.09\% R@10 on MMQA and 83.25\% R@2 on WebQA. Notably, relative improvements are largest on weaker retrievers such as Vis-BGE-m3, while top tier systems like InternVL-G still yield non-trivial gains.

\begin{figure}[t]
    \centering
    \includegraphics[width=1\linewidth]{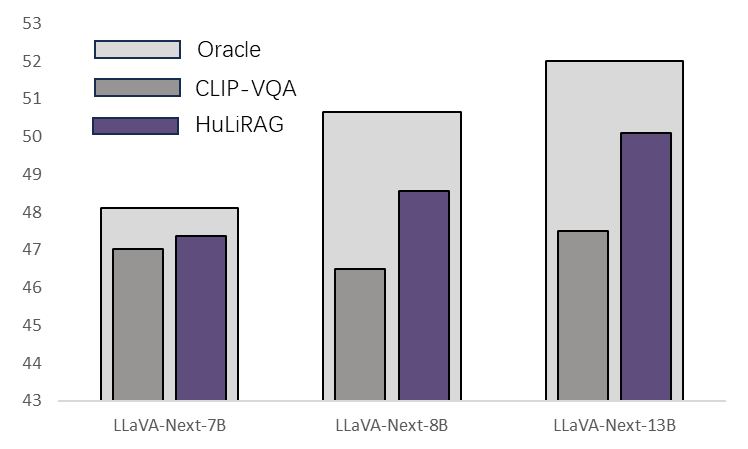}
    \caption{Evaluating LLaVA-Next (7B/8B/13B) with CLIP$\rightarrow$VQA, HuLiRAG, and Oracle under the LLM-as-a-Judge protocol.}
    \label{fig:llava-judge}
\end{figure}


\begin{figure}[ht]
\centering
\includegraphics[width=0.45\textwidth]{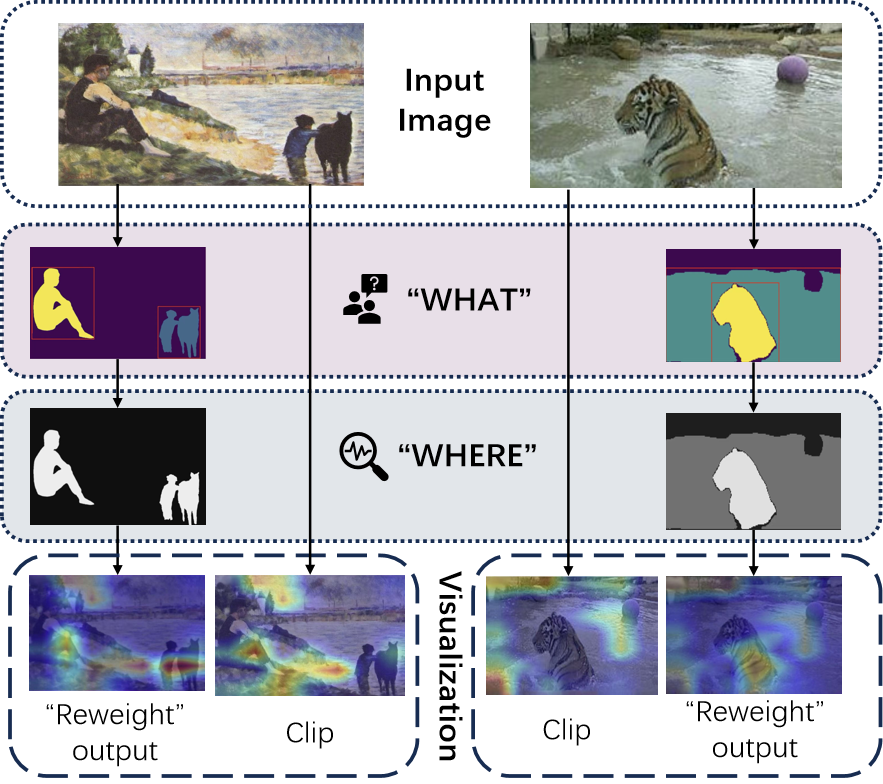}
\caption{
Comparison between CLIP’s global attention (query-agnostic) and HuLiRAG’s \textit{What-Where-Reweight} mechanism (query-conditioned). HuLiRAG mimics human perception by adaptively fusing global context with masked local regions. The HuLiRAG heatmaps are obtained by overlaying regional relevance on top of CLIP’s original activation, showing how adaptive fusion sharpens attention to query-relevant evidence.
}
\label{fig:visual_analysis}
\end{figure}

\begin{figure*}[t]
    \centering
    \includegraphics[width=\linewidth]{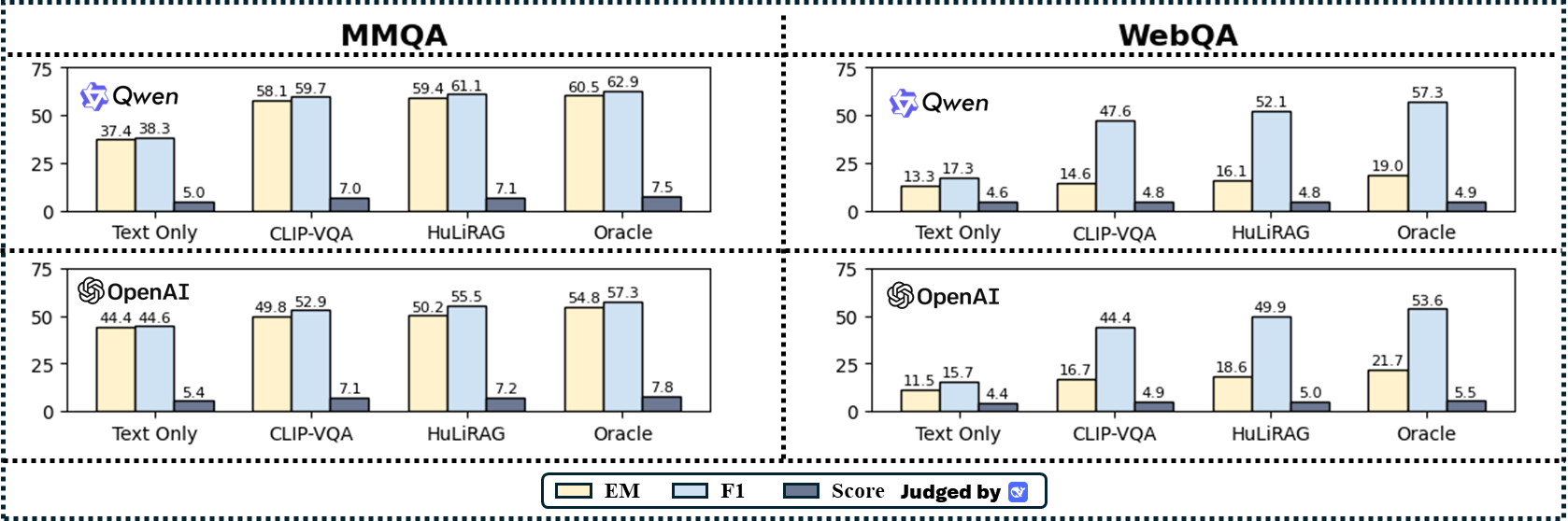}
    \caption{Evaluation by business LLM}
    \label{fig:placeholder}
\end{figure*}

\noindent\textbf{Main VQA Results.}
Table~\ref{tab:mq-webqa-final} shows that our HuLiRAG pipeline consistently improves performance across MMQA and WebQA. On MMQA, InternVL-1B improves from 30.00 to 41.14 EM with HuLiRAG, recovering 53\% of the gap toward its GT oracle (50.56), with larger backbones exhibiting comparable recovery ratios. Since MMQA primarily evaluates short, factoid-style answers, EM is the natural metric. On WebQA, which involves longer free-form responses, HuLiRAG consistently adds 3\%--6\% F1 over the CLIP$\rightarrow$VQA baseline, and the gains persist across scales. Within HuLiRAG, the proposed Mask-Guided FT yields an additional 0.4\%--3.5\% absolute improvement, indicating that domain-aware supervision not only strengthens retrieval gains but also mitigates instability. Although introduced here as part of HuLiRAG, this lightweight mechanism is transferable and can be seamlessly applied to fine-tuning other LLM backbones for VQA where region-aware signals are advantageous. These findings clarify that HuLiRAG equipped with Mask-Guided FT not only narrows the gap toward oracle performance but also provides a broadly applicable strategy for stabilizing and amplifying retrieval-based VQA.

To complement the InternVL results, we further benchmark the LLaVA-Next family (7B, 8B, 13B) under our retrieval settings. Unlike WebQA F1, which may conflate retrieval accuracy with scaling effects of large models, we adopt the LLM-as-a-Judge protocol \citep{zheng2023judging} based on MT-Bench and Chatbot Arena. Using the fixed single-answer grading prompt (shown in Table~\ref{tab:eval_system}), we employ GPT-4 as the judging model to score each response, and compare CLIP$\rightarrow$VQA, our HuLiRAG pipeline, and the GT-Image oracle. As shown in Fig.~\ref{fig:llava-judge}, HuLiRAG consistently improves over CLIP$\rightarrow$VQA across all scales, closely tracking the oracle even at 7B, and narrowing the gap further at 8B and 13B (e.g., on LLaVA-Next-13B HuLiRAG reduces more than half of the gap to the oracle).

We further evaluate with Qwen \cite{bai2023qwen} and GPT4o\cite{achiam2023gpt} as answer generators, while DeepSeek \cite{liu2024deepseek} serves as the judge. In addition to EM and F1, DeepSeek assigns a holistic Score on a 0--10 scale, following a rubric that weighs helpfulness, accuracy, depth, and clarity. The detailed criteria for this scoring protocol are summarized in Table~\ref{tab:eval_rubric_long}.
As shown in Figure~\ref{fig:placeholder}, HuLiRAG consistently raises judged quality across MMQA and WebQA, moving closer to the oracle. 
For instance, on MMQA with Qwen as the generator, HuLiRAG improves the judged Score from 6.98 (CLIP$\rightarrow$VQA) to 7.11, approaching the oracle level of 7.45.

\subsection{Analysis}

\textbf{Refinement Study: Effect of Fusion Strategies.} 
Table~\ref{tab:rerank-fixed} reports the effect of different similarity fusion strategies in HuLiRAG-Ret. 
Global and local levels serve as reference points that rely solely on holistic or region-aware similarity without any fusion. 
All fusion strategies yield clear improvements, confirming the benefit of combining the two. 
Addition performs simple linear blending, offering moderate gains (MMQA R@1: 85.83→86.04). 
Multiply further enhances precision by emphasizing regions where both cues are strong (MMQA R@1: 86.19, WebQA R@2: 70.50). 
Learnable Reweight achieves the best overall performance (MMQA R@1: 86.71, WebQA R@2: 71.38, WebQA R@10: 86.64), 
as it adaptively calibrates the contributions of global and local similarity through a lightweight parameterized fusion. 
These results highlight that adaptive weighting provides the most balanced and discriminative retrieval alignment.


\begin{table}[t]
\centering
\scriptsize
\caption{
Ablation of HuLiRAG components on MMQA and WebQA.
}
\label{tab:component}
\resizebox{0.9\linewidth}{!}{%
\begin{tabular}{lcc}
\toprule
Variant & MMQA R@1 & WebQA F1 \\
\midrule
\cellcolor{slategray}Full HuLiRAG & \cellcolor{slategray}87.57 & \cellcolor{slategray}76.79 \\
w/o What & 70.38 & 63.52 \\
w/o Where & 79.53 & 70.16 \\
w/o Reweight & 86.04 & 75.15 \\
\bottomrule
\end{tabular}
}
\end{table}

\textbf{Visual Cognitive Pipeline: Emulating Expert Reasoning.}
For queries requiring multi-layered understanding (e.g., “What kind of animal is close to a human in Seurat’s painting and \textit{Portrait of a Young Nobleman}?”), our architecture follows a three-stage pipeline (Figure~\ref{fig:visual_analysis}) that surpasses conventional models trapped by global feature homogenization. The “WHAT” stage localizes entities using open-vocabulary prompts and region masks, for instance localizing the child and dog in Seurat’s painting while minimizing the influence of the seated figure. The “WHERE” stage then disentangles spatial relations via binary masks, enabling fine-grained reasoning such as separating a tiger’s contour from rippling water where CLIP conflates “tiger” with “aquatic context.” Finally, the “Reweight” stage shifts attention toward interaction zones, concentrating on task-relevant regions like the tiger’s body, while CLIP disperses activation over distractors. This pipeline, instantiated in HuLiRAG’s design, systematically integrates global context with localized evidence, ensuring reasoning that is both semantically faithful and spatially grounded. By emulating expert workflows that begin with global composition and proceed to fine-grained inspection, the framework avoids shallow feature matching and instead produces structured, query-relevant reasoning that generalizes across diverse domains.

\textbf{Component Analysis.}We further analyze the contributions of HuLiRAG’s stages in Table 4.
The \textit{what} stage is essential because it anchors free-form textual queries to visual referents and decomposes them into grounded entities that guide SAM and Alpha-CLIP for localized alignment.
Without it, the model feeds the entire sentence directly to SAM for segmentation, producing coarse and semantically inconsistent regions that are later encoded by Alpha-CLIP, leading to severe semantic drift (MMQA R@1: 72.1, WebQA F1: 65.1).
Removing the \textit{where} stage disables spatial grounding by replacing GroundingDINO and SAM with a full-image mask, collapsing reranking into a global comparison between CLIP and Alpha-CLIP embeddings and causing a notable drop in precision (MMQA R@1: 79.5, WebQA F1: 70.1).
Discarding the \textit{reweight} stage retains both global and local cues but fuses them linearly, reducing adaptivity and discriminative strength (MMQA R@1: 86.0, WebQA F1: 75.2). In addition to the three stages, mask-guided fine-tuning, though introduced as part of HuLiRAG, constitutes a general mechanism for enforcing spatial grounding: across scales, it consistently improves retrieval-augmented VQA (e.g., InternVL-2B EM: 65.0$\rightarrow$69.0; WebQA F1: 74.9$\rightarrow$77.6), underscoring both its robustness and generality.

\section{Conclusion}
In this work, we introduced Human-Like Retrieval-Augmented Generation (HuLiRAG), a cognitively inspired framework that overcomes static global-only retrieval. Through a \textit{what–where–reweight} mechanism, HuLiRAG grounds queries in fine-grained visual evidence by integrating objects, relations, and semantics. This design reflects human perceptual reasoning, adaptively balancing global context with local details. Experiments on WebQA and MultimodalQA show consistent gains in retrieval precision and answer consistency, underscoring that cognitively motivated architectures can improve both interpretability and performance in multimodal reasoning.

\balance


\bibliographystyle{ACM-Reference-Format}
\bibliography{acmart.bib}

\appendix

\begin{table*}[t]
\centering
\caption{Evaluation instructions for the judging system.}
\label{tab:eval_system}
\renewcommand{\arraystretch}{1.3}
\small
\begin{tabular}{p{0.25\linewidth}p{0.72\linewidth}}
\toprule
\rowcolor[HTML]{F2F2F2}
\multicolumn{2}{c}{\textbf{System Instruction}} \\ 
\midrule
\textbf{Prompt} & Please act as an impartial judge and evaluate the quality of the response provided by an AI assistant to the user question displayed below. Your evaluation should consider factors such as the helpfulness, relevance, accuracy, depth, creativity, and level of detail of the response. \newline
Begin your evaluation by providing a short explanation. Be as objective as possible. After providing your explanation, please rate the response on a scale of \textbf{1 to 100}, where only integer scores are allowed, by strictly following this format: \texttt{"Rating: [[X]]"}, for example: \texttt{"Rating: [[85]]"}. \\

\midrule
\textbf{Format} & 
\textbf{[Question]} \newline
\texttt{\{question\}} \newline
\textbf{[The Start of Assistant’s Answer]} \newline
\texttt{\{answer\}} \newline
\textbf{[The End of Assistant’s Answer]} \\
\bottomrule
\end{tabular}
\end{table*}

\begin{table*}[t]
\centering
\caption{Extended evaluation rubric showing detailed criteria (Helpfulness, Accuracy, Depth, Clarity) with weights, ranges, and scoring rules.}
\label{tab:eval_rubric_long}
\renewcommand{\arraystretch}{1.4}
\small
\begin{tabular}{p{0.25\linewidth}p{0.72\linewidth}}
\toprule
\rowcolor[HTML]{F2F2F2}
\multicolumn{2}{c}{\textbf{Evaluation Criteria (1--10 scale, weighted)}} \\ 
\midrule
\textbf{Helpfulness (0.35)} & Measures whether the answer resolves the user’s intent with actionable, directly useful content. \newline 
\textbf{1--3}: Misses intent or irrelevant. \newline 
\textbf{4--6}: Partial resolution with significant gaps. \newline 
\textbf{7--8}: Good coverage, minor omissions. \newline 
\textbf{9--10}: Fully actionable, comprehensive, anticipates implicit needs. \\
\midrule
\textbf{Accuracy (0.35)} & Captures factual correctness and avoidance of misleading claims. \newline 
\textbf{1--3}: Major errors or hallucinations. \newline 
\textbf{4--6}: Noticeable inaccuracies or overgeneralizations. \newline 
\textbf{7--8}: Mostly accurate with minor issues. \newline 
\textbf{9--10}: Perfectly accurate, with explicit handling of uncertainty. \\
\midrule
\textbf{Depth (0.20)} & Evaluates reasoning structure and completeness. \newline 
\textbf{1--3}: Extremely superficial, no reasoning. \newline 
\textbf{4--6}: Minimal reasoning, key steps missing. \newline 
\textbf{7--8}: Solid explanation, but shallow in scope. \newline 
\textbf{9--10}: Thorough, structured, anticipates edge cases and alternative views. \\
\midrule
\textbf{Clarity (0.10)} & Focuses on readability, organization, and precision. \newline 
\textbf{1--3}: Unintelligible or incoherent. \newline 
\textbf{4--6}: Hard to follow due to verbosity or poor structure. \newline 
\textbf{7--8}: Understandable but minor awkwardness. \newline 
\textbf{9--10}: Polished, concise, professional academic tone. \\
\midrule
\rowcolor[HTML]{F2F2F2}
\multicolumn{2}{c}{\textbf{Scoring Protocol}} \\
\midrule
\textbf{Penalties} & Safety violations, hallucinations, fabricated citations, ignoring constraints, verbosity, or incomplete answers incur \textbf{--2 to --10} deductions depending on severity. \\
\midrule
\textbf{Bonuses} & Exceptional uncertainty handling (+0.5), perfect formatting (+0.3), highly relevant edge cases (+0.5). Bonuses are capped at 10.0. \\
\midrule
\textbf{Final Output} & Strictly four numbers: \texttt{Helpfulness Accuracy Depth Clarity}, each with \textbf{two decimals} (e.g., \texttt{8.50 9.25 7.00 9.00}). No extra text, JSON, or explanation is permitted. \\
\bottomrule
\end{tabular}
\end{table*}

\begin{table*}[t]
    \centering
    \caption{Case study examples from WebQA and MultimodalQA, showing how model improves answer grounding compared to text-only and CLIP-VQA baselines.}

    \small
    \begin{tabular}{p{0.25\columnwidth}p{1.75\columnwidth}}
        \toprule

        \rowcolor[HTML]{F2F2F2}
        \multicolumn{2}{c}{\textbf{Examples in WebQA}} 
        \\ \midrule 
        \multicolumn{2}{c}{\textbf{Example 1}} 
        \\ \midrule 
        \textbf{Question} & “What objects are next to the stanchions on the sidewalk in front of the Spice of Life?” \\
        \textbf{Ground Truth} & \textcolor[HTML]{B22222}{Phone booths} are next to the stanchions on the sidewalk in front of the Spice of Life. \\
        \textbf{Text-only} & \textcolor[HTML]{B22222}{Street signs} are next to the stanchions. \\
        \textbf{CLIP-VQA} & \textcolor[HTML]{B22222}{Bus stops} are next to the stanchions. \\
        \textbf{HuLiRAG-VQA} & \textcolor[HTML]{B22222}{Phone booths} are next to the stanchions, along with a \textcolor[HTML]{B22222}{street sign}. \\ 
        \midrule
        \multicolumn{2}{c}{\textbf{Example 2}} 
        \\ \midrule 
        \textbf{Question} & “What hangs in the covered section of Via Amore off Worth Avenue in Palm Beach, Florida?” \\
        \textbf{Ground Truth} & A \textcolor[HTML]{B22222}{lantern} hangs in the covered section of Via Amore off Worth Avenue in Palm Beach, Florida. \\
        \textbf{Text-only} & There is a hanging \textcolor[HTML]{B22222}{shop sign} or possibly a \textcolor[HTML]{B22222}{banner} in the covered walkway, which indicates the presence of a store or restaurant. \\
        \textbf{CLIP-VQA} & A decorative \textcolor[HTML]{B22222}{light fixture} or \textcolor[HTML]{B22222}{chandelier} is hanging in the covered section, providing illumination for the walkway. \\
        \textbf{HuLiRAG-VQA} & A \textcolor[HTML]{B22222}{lantern} is hanging from the ceiling in the covered section, and it appears alongside some decorative elements such as a small \textcolor[HTML]{B22222}{sign} or \textcolor[HTML]{B22222}{plant}. \\ 
        \midrule
        \multicolumn{2}{c}{\textbf{Example 3}} 
        \\ \midrule 
        \textbf{Question} & “What tops did the women wear for the discus throw competition at the 2018 Finnish Athletics Championships?” \\
        \textbf{Ground Truth} & Women wore \textcolor[HTML]{B22222}{t-shirts} and \textcolor[HTML]{B22222}{tank tops} at the discus throw competition during the 2018 Finnish Athletics Championships. \\
        \textbf{Text-only} & The women were wearing athletic uniforms, such as \textcolor[HTML]{B22222}{jerseys} and \textcolor[HTML]{B22222}{sports bras}, which are typical for track and field events. \\
        \textbf{CLIP-VQA} & Most of the women wore sleeveless athletic tops or \textcolor[HTML]{B22222}{sports bras}, while some had competition \textcolor[HTML]{B22222}{singlets}. \\
        \textbf{HuLiRAG-VQA} & The women can be seen wearing a mix of \textcolor[HTML]{B22222}{t-shirts} and sleeveless tops, such as \textcolor[HTML]{B22222}{tank tops}, with a few athletes also in lighter athletic \textcolor[HTML]{B22222}{vests}. \\
        \midrule 
        \multicolumn{2}{c}{\textbf{Example 4}} 
        \\ \midrule 
        \textbf{Question} & “Which organ's pipes are more golden in color; St Georges Church Organ, Hanover Square, London or Ponce Cathedral pipe organ?” \\
        \textbf{Ground Truth} & \textcolor[HTML]{B22222}{St Georges Church Organ} is more golden in color compared to the Ponce Cathedral pipe organ. \\
        \textbf{Text-only} & The \textcolor[HTML]{B22222}{Ponce Cathedral pipe organ} is often described as ornate and golden in appearance, suggesting it is more golden than the one in London. \\
        \textbf{CLIP-VQA} & The pipes of the \textcolor[HTML]{B22222}{St Georges Church Organ} appear more golden in the image compared to those of the Ponce Cathedral organ. \\
        \textbf{HuLiRAG-VQA} & The \textcolor[HTML]{B22222}{St Georges Church Organ} pipes look more golden overall, though the Ponce Cathedral organ also has metallic highlights. \\

         \bottomrule

         \rowcolor[HTML]{F2F2F2}
        \multicolumn{2}{c}{\textbf{Examples in MultimodalQA}} 
        \\ \midrule 
        \multicolumn{2}{c}{\textbf{Example 1}} 
        \\ \midrule 
        \textbf{Question} & “How many colors are on the Mississippi flag?” \\
        \textbf{Ground Truth} & \textcolor[HTML]{B22222}{3} (red, white, blue). \\
        \textbf{Text-only} & The Mississippi state flag traditionally has \textcolor[HTML]{B22222}{three} main colors, red, white, and blue, though some descriptions also mention \textcolor[HTML]{B22222}{gold}. \\
        \textbf{CLIP-VQA} & The flag appears to contain \textcolor[HTML]{B22222}{red, white, blue, and gold}, making it look like four colors in total. \\
        \textbf{HuLiRAG-VQA} & The Mississippi flag shows \textcolor[HTML]{B22222}{three} clear colors—red, white, and blue—although parts of the emblem may look golden. \\ 
        \midrule
        \multicolumn{2}{c}{\textbf{Example 2}} 
        \\ \midrule 
        \textbf{Question} & “Is Glen E. Friedman's jacket zipped or unzipped?” \\
        \textbf{Ground Truth} & The jacket is \textcolor[HTML]{B22222}{unzipped}. \\
        \textbf{Text-only} & It looks like the jacket is usually worn \textcolor[HTML]{B22222}{zipped}, as most portraits show it closed. \\
        \textbf{CLIP-VQA} & The jacket appears to be \textcolor[HTML]{B22222}{zipped}, since the front looks mostly closed. \\
        \textbf{HuLiRAG-VQA} & The jacket is \textcolor[HTML]{B22222}{unzipped}, with the front open and the zipper not fastened, revealing the clothing underneath. \\ 
        \midrule
        \multicolumn{2}{c}{\textbf{Example 3}} 
        \\ \midrule 
        \textbf{Question} & “What is hitting the bongos in the Incredible Bongo Band cover?” \\
        \textbf{Ground Truth} & \textcolor[HTML]{B22222}{Hands} are hitting the bongos. \\
        \textbf{Text-only} & The bongos are being played with \textcolor[HTML]{B22222}{drumsticks} or mallets, which are often shown in band artwork. \\
        \textbf{CLIP-VQA} & The bongos are struck by \textcolor[HTML]{B22222}{arms}, showing a performer energetically playing them. \\
        \textbf{HuLiRAG-VQA} & The bongos are struck by human \textcolor[HTML]{B22222}{hands}, with both palms clearly shown on the cover. \\

        \multicolumn{2}{c}{\textbf{Example 4}} 
        \\ \midrule 
        \textbf{Question} & “What type of ball is the center of the FC Bolosani logo?” \\
        \textbf{Ground Truth} & A \textcolor[HTML]{B22222}{soccer} ball is at the center of the FC Bolosani logo. \\
        \textbf{Text-only} & The logo likely has a \textcolor[HTML]{B22222}{basketball} in the center, as many European sports clubs are associated with basketball teams, and the round design with simple stripes suggests a basketball rather than a soccer ball. \\
        \textbf{CLIP-VQA} & The middle of the emblem looks like it contains a \textcolor[HTML]{B22222}{volleyball}-style object, with curved lines across the surface, giving the impression of a volleyball rather than a traditional soccer ball that has hexagonal patches. \\
        \textbf{HuLiRAG-VQA} & At the center of the FC Bolosani crest, there is clearly a \textcolor[HTML]{B22222}{soccer} ball, with visible hexagonal and pentagonal panels that confirm its design, although some shading and stylization of the logo could make it look slightly different at first glance. \\

        \bottomrule
    \end{tabular}
    \label{tab:case1}
\end{table*}

\end{document}